\documentclass[preprint,12pt]{elsarticle}

\usepackage{lineno}
\usepackage{graphicx}
\usepackage{rotating}
\usepackage{verbatim}%multiple line comment
\usepackage{amsmath,  amssymb}
\usepackage{subfigure}
\usepackage{algorithm}
\usepackage{algorithmic}
\usepackage{lscape}
\begin{document}
%\setpagewiselinenumbers
%\modulolinenumbers[1]
%\linenumbers

\title{A Time Series Forest for \\
Classification and Feature Extraction\tnoteref{t2}}

\author{Houtao Deng}\corref{cor1}
\ead{hdeng3@asu.edu}
\address{Intuit, Mountain View, CA, USA}
\author{George Runger}
\ead{george.runger@asu.edu}
\address{Arizona State University, Tempe, AZ, USA}
\author{Eugene Tuv}
\ead{eugene.tuv@intel.com}
\author{Martyanov Vladimir}
\ead{vladimir.martyanov@intel.com}
\address{Intel, Chandler, AZ, USA}
\cortext[cor1]{Corresponding author: hdeng3@asu.edu}

\begin{abstract}
A tree-ensemble method, referred to as time series forest (TSF), is proposed for time series classification. TSF employs a combination of entropy gain and a distance measure, referred to as the Entrance (entropy and distance) gain, for evaluating the splits. Experimental studies show that the Entrance gain improves the accuracy of TSF. TSF randomly samples features at each tree node and has computational complexity linear in the length of time series, and can be built using parallel computing techniques. The temporal importance curve is proposed to capture the temporal characteristics useful for classification. Experimental studies show that TSF using simple features such as mean, standard deviation and slope is computationally efficient and outperforms strong competitors such as one-nearest-neighbor classifiers with dynamic time warping. %, and the temporal importance curves can provide insights into the temporal characteristics.
\end{abstract}


\maketitle

 \newtheorem{defn}{Definition}

\section{Introduction}\label{section:introduction}
Time series classification has been playing an important role in many disciplines such as finance \citep{zeng2008supervised} and medicine \citep{costa2009constrained}. Although one can treat the value of each time point as a feature and use a regular classifier such as one-nearest-neighbor (NN) with Euclidean distance for time series classification, the classifier may be sensitive to the distortion of the time axis and can lead to unsatisfactory accuracy performance. One-nearest-neighbor with dynamic time warping (NNDTW) is robust to the distortion of the time axis and has proven exceptionally difficult to beat \cite{xi2006fast}. However, NNDTW provides limited insights into the temporal characteristics useful for distinguishing time series from different classes.

\begin{figure}[!h]
\centering
    \includegraphics[width=4 in,height=3 in]{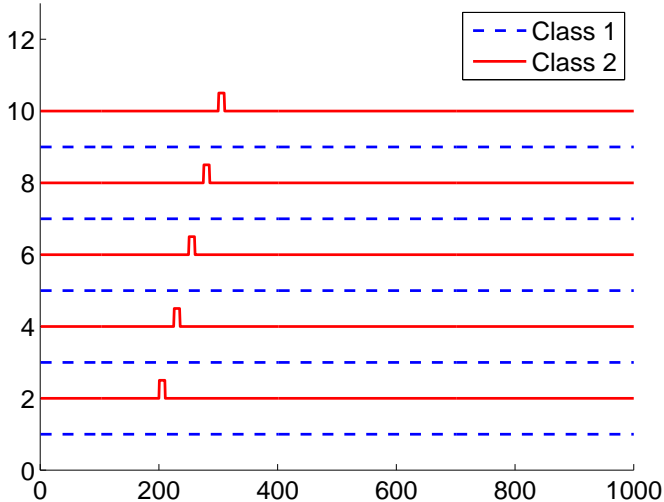}
\caption{\label{fig:demoInterval} The time series from class 2 have sudden changes between time 201 and time 400. An interval feature such as the standard deviation between time 201 and time 400 can distinguish the time series from the two classes.}
\end{figure}

The temporal features calculated over time series intervals \citep{rodriguez2001boosting}, referred to as interval features, can capture the temporal characteristics, and can also handle the distortion in the time axis. For example, in the two-class time series shown in Figure \ref{fig:demoInterval}, the time series from one of the classes have sudden changes between time 201 and time 400 but not in the same time points. An interval feature such as the standard deviation between time 201 and time 400 is able to distinguish the two-class time series.

\def\myWidth{2.3}

Previous work \citep{rodriguez2001boosting} has built decision trees on interval features. However, a large number of interval features can be extracted from time series, and there can be a large number of candidate splits to evaluate at each tree node. Class-based measures (e.g., entropy gain), which evaluate the ability of separating the classes, are commonly used to select the best split in a node. However, there can be many splits having the same ability of separating the classes. Therefore, measures able to further distinguish these splits are desirable. Also, given a large number of features/splits, an efficient and accurate classifier that can provide insights into the temporal characteristics is valuable.

To this end, we propose a novel tree-ensemble classifier: time series forest (TSF). TSF employs a new measure called the \emph{Entrance} (entropy and distance) gain to identify high-quality splits. We show that TSF using Entrance gain outperforms TSF using entropy gain and also two NNDTW algorithms. By using a random feature sampling strategy, TSF has computational complexity linear in the time series length.
%has a computational complexity linear to the number of time values, i.e., $O(AM)$, where $A$ is the computational complexity for evaluating one feature and $M$ is the time series length.
Furthermore, we propose the temporal importance curve to capture the temporal characteristics informative for time series classification.

The remainder of this paper is organized as follows.
%Section \ref{sec:background} provides background knowledge for the sections following it.
Section \ref{sec:related} presents the definition of the problem and related work. Section \ref{sec:Interval} introduces the interval features. Section \ref{sec:RF} describes the TSF method. Section \ref{sec:exp} demonstrates the effectiveness and efficiency of TSF by experiments. Conclusions are drawn in Section \ref{sec:conclusions}.

\section{Definition and Related Work}\label{sec:related}
Given $N$ training time series instances (examples): $\{e_1,..., e_i,...,e_N\}$ and the corresponding class labels $\{y_1,..., y_i,...,y_N\}$, where $y_i \in \{1,2,...,C\}$, the objective of time series classification is to predict the class labels for testing instances. Here we assume the values of time series are measured at equally-spaced intervals, and also assume the training and testing time series instances are of the same length $M$.

Time series classification methods can be divided into instance-based and feature-based methods. Instance-based classifiers predict a testing instance based on its similarity to the training instances. Among instance-based classifiers, nearest-neighbor classifiers with Euclidean distance (NNEuclidean) or dynamic time warping (NNDTW) have been widely and successfully used \citep{ratanamahatana2004making,xing2009early,jeong2011weighted,yu2011dynamic}. Usually NNDTW performs better than NNEuclidean (dynamic time warping \citep{Sakoe78dynamicprogramming} is robust to the distortion in the time axis), and is considered as a strong solution for time series problems \citep{ratanamahatana2005three}. Instance-based classifiers can be accurate, but they provide limited insights into the temporal characteristics useful for classification. %lack of interpretability is one disadvantage because on the training instances. %testing time are two major problems because there is no model built on the training data. %, A disadvantage of NNDTW is lack of interpretability.
%\citet{Yamada03decision-treeinduction} proposed a tree-based approach that splits instances according to a dissimilarity measure based on dynamic time warping. Such a decision tree is more interpretable than NNDTW, but the interpretability is still limited.

Feature-based classifiers build models on temporal features, and potentially can be more interpretable than instance-based classifiers. Feature-based classifiers commonly consist of two steps: defining the temporal features and training a classifier based on the temporal features defined.
\citet{Nanopoulos01feature} extracted statistical features such as the mean and deviation of an entire time series, and then used a multi-layer perceptron neural network for classification. This method only captured the global properties of time series. Local properties, potentially informative for classification, were ignored. \citet{geurts2001pattern} extracted local temporal properties after discretizing the time series. \citet{rodriguez2001boosting} boosted binary stumps on temporal features from intervals of the time series and \citet{rodriguez2004interval,rodrguez2005support} applied classifiers such as a decision tree and a SVM on the temporal features extracted from the boosted binary stumps.
However, only binary stumps were boosted, and the effect of using more complex base learners, such as decision trees, should be studied \citep{rodriguez2001boosting} (but larger tree models impact the computational complexity).
Furthermore, in decision trees \citep{rodriguez2001boosting,rodriguez2004interval,rodrguez2005support} class-based measures are often used to evaluate the candidate splits in a node. However, the number of candidate splits is generally large, and, thus, there can be multiple splits having the same ability of separating the classes. Consequently, additional measures able to further distinguish these features are desirable. \citet{ye2009time} briefly discussed strategies of introducing additional measures to break ties, but it was in a different context.

Recently, \citet{ye2009time} proposed time series shapelets to perform interpretable time series classification. Shapelets are time series subsequences which are in some sense maximally representative of a class \citep{ye2009time}. \citet{ye2009time,Xing2011,Lines2012} have successfully shown that time series shapelets can produce highly interpretable results. In term of accuracy, \citet{Lines2012} showed that the shapelet approach is comparable to NNDTW for nine data sets investigated.

\citet{eruhimov2007} considered a massive number of features. The feature sets were derived from statistical moments, wavelets, Chebyshev coefficients, PCA coefficients, and the original values of time series.
The method can be accurate, but is hard to interpret and computationally expensive. The objective of our work is to produce an effective and efficient classifier that uses/yields a set of simple features that can contribute to the domain knowledge. For example, in manufacturing applications, specific properties of the time series signals that discriminate conforming from un-conforming products are invaluable to diagnose, correct, and improve processes.

\section{Interval Features \label{sec:Interval}}
Interval features are calculated from a time series interval, e.g., ``the interval between time 10 and time 30". Many types of features over a time interval can be considered, but one may prefer simple and interpretable features such as the mean and standard deviation, e.g., ``the average of the time series segment between time 10 and time 30".

Let $K$ be the number of feature types and $f_k(\cdot)$ ($k=1,2,...,K$) be the $k^{th}$ type. Here we consider three types: $f_1=mean$, $f_2=standard\ deviation$, $f_3=slope$. Let $f_k(t_1,t_2)$ for $1 \leq t_1 \leq t_2 \leq M$ denote the $k^{th}$ interval feature calculated over the interval between $t_1$ and $t_2$. Let $v_i$ be the value at time $i$ for a time series example. Then the three interval features for the example are calculated as follows:

 \begin{equation}\label{meanFormula}
f_1(t_1,t_2) = \frac{\sum_{i=t_1}^{t_2}v_i}{t_2-t_1+1}
\end{equation}
\begin{equation}
f_2(t_1,t_2)  =
\begin{cases}
 \sqrt{\frac{\sum_{i=t_1}^{t_2}{(v_i - f_1(t_1,t_2))}^2}{t_2-t_1}} & \text{$t_2>t_1$} \\
     \ \ \ \ \ \    0 & \text{$t_2 = t_1$} \\
\end{cases}
\end{equation}
\begin{equation}
f_3(t_1,t_2)  =
\begin{cases}
  \hat \beta & \text{$t_2>t_1$} \\
        0 & \text{$t_2 = t_1$} \\
\end{cases}
\end{equation}
where $\hat \beta$ is the slope of the least squares regression line of the training set $\{(t_1,v_{t_1})$, $(t_1+1,v_{t_1+1}). \ldots, (t_2,v_{t_2})\}$.%when $t_2>t_1$; and 0 when $t_2 = t_1$.

Interval features have been shown to be effective for time series classification \citep{rodriguez2001boosting,rodriguez2004interval,rodrguez2005support}. However, the interval feature space is large ($O(M^2)$). \citet{rodriguez2001boosting} considered using only intervals of lengths equal to powers of two, and, therefore, reduced the feature space to $O(M \log M)$. Here we consider the random sampling strategy used in a random forest \citep{breiman2001} that reduces the feature space to $O(M)$ at each tree node. %In addition, we propose a novel splitting criterion that considers both entropy gain and a distance measure.

\section{Time Series Forest Classifier \label{sec:RF}}
\subsection{Splitting criterion}
A time series tree is the base component of a time series forest, and the splitting criterion is used to determine the best way to split a node in a tree.
%Such a tree uses an interval feature for splitting the training instances and uses an evaluation criterion combining the entropy gain and distance measures at each node.
A candidate split $S$ in a time series tree node tests the following condition (for simplicity and without loss of generality, we assume the root node here):
 \begin{equation}\label{split}
f_k(t_1,t_2) \le \tau
\end{equation}
for a threshold $\tau$. The instances satisfying the condition are sent to the left child node. Otherwise, the instances are sent to the right child node.

Let $\{f_k^n(t_1,t_2), n\in {1,2,...,N}\}$ denote the set of values of $f_k(t_1,t_2)$ for all training instances at the node. To obtain a good threshold $\tau$ in equation \ref{split}, one can sort
the feature values of all the training instances and then select the best threshold from the midpoints between pairs of consecutive values, but this can be too costly \citep{rodriguez2004interval}.
%one can sort the instances based on their $f_k(t_1,t_2)$ value, and then test the average of $f_k(t_1,t_2)$ from two consecutive instances like C4.5 \citet{quinlan1993c4}. one can evaluate each of the $$ value from this set similar to C4.5 \citep{quinlan1993c4}, which has $O(N)$ tests for selecting the best threshold.
We consider the strategy employed in \citet{rodriguez2004interval}.
 %The thresholds are selected in a way that the range of values of the feature is divided in uniform-width intervals.
The candidate thresholds for a particular type feature $f_k$ are formed such that the range of $[\min_{n=1}^N(f_k^n(t_1,t_2))$, $\max_{n=1}^N(f_k^n(t_1,t_2)]$ is divided into equal-width intervals. The number of candidate thresholds is denoted as $\kappa$ and is fixed, e.g., 20. The best threshold is then selected from the candidate thresholds. In this manner, sorting is avoided, and only $\kappa$ tests are needed. %the computational complexity for searching a threshold is

Furthermore, a splitting criterion is needed to define the best split $S^*$: $f_*(t_1^*,t_2^*) \le \tau^*$. We employ a combination of entropy gain and a distance measure as the splitting criterion.
Entropy gain are commonly used as the splitting criterion in tree models. Denote the proportions of instances corresponding to classes $\{1,2,...,C\}$ at a tree node as $\{\gamma_1,\gamma_2,...,\gamma_C\}$, respectively. The entropy at the node is defined as
\begin{equation}\label{eq:entropy}
Entropy = -\Sigma_{c=1}^{C}\gamma_c\log \gamma_c
\end{equation}
The entropy gain \textbf{$\bigtriangleup Entropy$} for a split is then the difference between the weighted sum of entropy at the child nodes and the entropy at the parent node, where the weight at a child node is the proportion of instances assigned to that child node.

$\bigtriangleup Entropy$ evaluates the usefulness of separating the classes. However, in time series classification, the number of candidate splits can be large, and there are often cases where multiple candidate splits have the same $\bigtriangleup Entropy$. Therefore we consider an additional measure called $Margin$, which calculates the distance between a candidate threshold and its nearest feature value. The $Margin$ of split $f_k(t_1,t_2) \le \tau$ is calculated as
\begin{equation}
Margin = \min_{n=1,2,\ldots, N}|f_k^n(t_1,t_2)-\tau|
\end{equation}
where $f_k^n(t_1,t_2)$ is the value of $f_k(t_1,t_2)$ for the $n^{th}$ instance at the node. %A larger $Margin$ is preferred.
%We only consider splits over $|threshold|$ equally spaced thresholds. Therefore, it is useful to use the $Margin$ as an additional criterion to select the best split.% and interpretability.
A new splitting criterion $E$, referred to as the \emph{Entrance} (entropy and distance) gain, is defined as a combination of $\bigtriangleup Entropy$ and $Margin$.
\begin{equation}\label{eq:evaluation}
E = \bigtriangleup Entropy + \alpha \cdot Margin
\end{equation}
where $\alpha$ is small enough so that the only role for $\alpha$ in the model is to break ties that can occur from the entropy gain alone. %To implement the Entrance gain in a tree-based algorithm, one can assign a very small value to $\alpha$.
Alternatively, one can store the values of $\bigtriangleup Entropy$ and $Margin$ for a split, and use $Margin$ to break ties when another split has the same $\bigtriangleup Entropy$.

Clearly, the split with the maximum $E$ should be selected to split the node. Furthermore, $Margin$ and $E$ are sensitive to the scale of the features, and we employ the following strategy if different types of features have different scales. For each feature type $f_k$, select the split with the maximum Entrance gain. To compare the best splits from different feature types, the split with the maximum $\bigtriangleup Entropy$ is selected. If the best splits from different feature types have the same maximum $\bigtriangleup Entropy$, one of the best splits is randomly selected.

\def\myWidth{3}
\begin{figure}[!h]
\centering
    \includegraphics[width= \myWidth in]{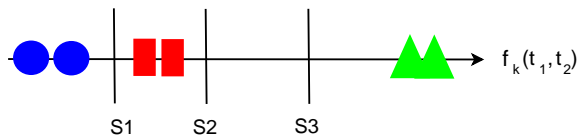}
\caption{Here the $x$-axis represents the value of an interval feature. The figure shows six instances associated with three classes (blue, red, and green), and three splits ($S_1$, $S_2$, and $S_3$) producing the same entropy gain. The Entrance gain $E$ is able to select $S_3$ as the best split. \label{fig:split}}
\end{figure}

Figure \ref{fig:split} illustrates the intuition behind the criterion $E$. The figure shows, in one dimension, six instances from three classes in different symbols/colors.
%Note the blue instances (filled circle) and the red instances (filled rectangle) are closer to each other than the green instances (filled triangle), and, thus, may be considered more similar to each other.
Three candidate splits $S_1$, $S_2$ and $S_3$ are also shown in the figure. Clearly, all splits have the same $\bigtriangleup Entropy$, but one may prefer $S_3$ because $S_3$ has a larger margin than $S_1$ and $S_2$. The Entrance gain is able to choose $S_3$ as the best split.

\begin{algorithm}[h!]
\small
\caption{$sample()$ function: randomly samples a set of intervals $<T_1,T_2>$, where $T_1$ is the set of starting time points of intervals, and $T_2$ is the set of ending points. The function $RandSampNoRep(set,samplesize)$ randomly selects $samplesize$ elements from $set$ without replacement.}
\label{alg:sampling}
\begin{algorithmic}
 \STATE $T_1=\emptyset$, $T_2=\emptyset$
\STATE $W = RandSampNoRep(\{1,..., M\}, \sqrt M)$
\FOR{$w$ in set $W$}
\STATE $T_1 = RandSampNoRep(\{1,...,M-w+1\}, \sqrt{M-w+1})$
\FOR{$t_1$ in set $T_1$}
 \STATE $T_2=T_2 \bigcup \ (t_1+w-1)$
\ENDFOR
\ENDFOR
\STATE return $<T_1,T_2>$
\end{algorithmic}
\end{algorithm}

\begin{algorithm}[h!]
%\scriptsize
\small
\caption{$tree(data)$: Time series tree. For simplicity of the algorithm, we assume different types of features are on the same scale so that $E$ can be compared.} % The function $evaluate(f_k(t_1,t_2) \leq \tau,data)$ calculates the criterion in equation \ref{eq:evaluation} on $data$.}
\label{treeAlg}
\begin{algorithmic}
 \STATE $<T_1,T_2>$=$sample()$
 \STATE calculate $Threshold_k$, the set of candidate thresholds for each feature type $k$
\STATE $E^* = 0$, $\bigtriangleup Entropy^* = 0$, $t_1^* = 0$, ${t_2}^* = 0$, $\tau^* = 0$, ${f}_* = \emptyset$
\FOR{$<t_1,t_2>$ in set $<T_1,T_2>$}
%\FOR{$f_i(\cdot)$ in set $f(\cdot)$}
\FOR{$k$ in 1:$K$}
\FOR{$\tau$ in $Threshold_k$}
 \STATE calculate $\bigtriangleup Entropy$ and $E$ for $f_k(t_1,t_2) \leq \tau$
\IF {$E > E^*$}
        \STATE $E^* = E$, $\bigtriangleup Entropy^* = \bigtriangleup Entropy$, $t_1^* = {t_1}$, $t_2^* = t_2$, $\tau^* = \tau$, ${f}_* = f_k$
\ENDIF
\ENDFOR
\ENDFOR
\ENDFOR
\IF {$\bigtriangleup Entropy^* = 0$}
     \STATE label this node as a leaf and return
\ENDIF

\STATE $data_{left}$ $\leftarrow$ time series with $f_*(t_1^*,t_2^*)\leq \tau^*$
\STATE $data_{right}$ $\leftarrow$ time series with $f_*(t_1^*,t_2^*) >  \tau^*$
\STATE $tree(data_{left})$
\STATE $tree(data_{right})$
\end{algorithmic}
\end{algorithm}

\subsection{Time Series Tree and Time Series Forest}
%\subsection{Forest Construction}
The construction of a time series tree follows a top-down, recursive strategy similar to standard decision tree algorithms, but uses the Entrance gain as the splitting criterion. Furthermore, the random sampling strategy employed in random forest (RF) \citep{breiman2001} is considered here. At each node, RF only tests $\sqrt p$ features randomly sampled from the complete feature set consisting of $p$ features. In each time series tree node, we consider randomly sampling $O(\sqrt{M})$ interval sizes and $O(\sqrt{M})$ starting positions. Therefore, the feature space is reduced to only $O(M)$. The sampling algorithm is illustrated in Algorithm \ref{alg:sampling}.

The time series tree algorithm is shown in Algorithm \ref{treeAlg}. For simplicity, we assume different types of features are on the same scale so that $E$ can be compared. If different types of features have different scales, the previous mentioned strategy can be used, that is, for each feature type $f_k$, select the split with the maximum Entrance gain. To compare the best splits from different feature types, the split with the maximum $\bigtriangleup Entropy$ is selected. Furthermore, a node is labeled as a leaf if there is no improvement on the entropy gain (e.g. all features have the same value or all instances belong to the same class).  %If the best splits from different types of features have the same maximum $\bigtriangleup Entropy$, a split is randomly selected.

A time series forest (TSF) is a collection of time series trees. A TSF predicts a testing instance to be the majority class according to the votes from all time series trees. %The random  selection reduces the number of features to be evaluated from $O(p)$ to $O(\sqrt p)$ at a node. %(assume the depth of tree is $O(\log N)$ where $N$ is the number of instances). %Similar to RF, TSF evaluates only a fraction of the original feature space at each node. %This yields the same benefits (reduced variance and computations).
%For TSF,  % %According to the experiments shown in Section \ref{sec:exp},

\subsection{Computational Complexity}
Let $n^i_j$ denote the number of instances in the $j^{th}$ node at the $i^{th}$ depth in a time series tree. At each node, calculating the splitting criterion of a single interval feature has complexity $O(n^i_j\kappa)$, where $\kappa$ is the number of candidate thresholds. As $O(M)$ interval features are randomly selected for evaluation, the complexity for evaluating the features at a node is $O(n^i_jM \kappa)$. %Selecting the best feature out of $M$ has a complexity $M$.
As $\kappa$ is considered as a constant, the complexity at a node is  $O(n^i_jM)$.

The total number of instances at each depth is at most $N$ (i.e., $\sum_j{n^i_j} \leq N$). Therefore, at the $i^{th}$ depth in the tree, the complexity is $O(\sum_j{n^i_jM})$ $\le$ $O({NM})$. Assuming the maximum depth of a tree model is $O(\log N)$ \citep{witten2005data}, the complexity of a time series tree becomes $O(MN\log N)$. Therefore, the complexity of a TSF with $nTree$ time series trees is at most $O(nTreeMN \log N)$, linear in the length of time series.

\subsection{Temporal Importance Curve}
TSF consists of multiple trees and is difficult to understand. Here we propose the temporal importance curve to provide insights into time series classification. At each node of TSF, the entropy gain can be calculated for the interval feature used for splitting. For a time index in the time series, one can add the entropy gain of all the splits associated with the time index for a particular type of feature. %from a specific type of interval feature that include that time point. That is,
That is, for a feature type $f_k$, the importance score for time index $t$ can be calculated as
\begin{equation}%\label{eq:M}
Imp_k(t)= \sum_{t_1 \leq t \leq t_2, \nu \in SN} \bigtriangleup Entropy(f_k(t_1,t_2),\nu)
\end{equation}
where $SN$ is the set of split nodes in TSF, and $\bigtriangleup Entropy(f_k(t_1,t_2),\nu)$ is the entropy gain for feature $f_k(t_1,t_2)$ at node $\nu$. Note $\bigtriangleup Entropy(f_k(t_1,t_2),\nu)=0$ if $f_k(t_1,t_2)$ is not used for splitting node $\nu$. %that is, $\bigtriangleup Entropy(f_i(t_1,t_2),S_k)$ can be nonzero only when $f_i(t_1,t_2)$ is used for splitting $n_k$.
Furthermore, one temporal importance curve is generated for each feature type. Consequently, for the mean, standard deviation and slope features, we calculate the mean, standard deviation, and slope temporal importance curves, respectively.
%Then the \textbf{naive temporal importance curve} for feature type $f_i$ is the plot of $Imp_i^0(t)$ for $t$=\{1, 2, ... M\}, that is, all the time points in the time series.

To investigate the temporal importance curve, we simulated two data sets, each with 1000 time points and two classes. For the first data set the time series have the same distribution so that no feature is useful for separating the classes. The time series values from both classes are normally distributed with zero mean and unit variance. The time series and the importance curves from TSF using Entrance gain are shown in Figure \ref{fig:simuUnifImpCurveEntrance}. It can be seen that all curves have larger values in the middle.

Note that the number of intervals that include time index $t$ in a time series is
\begin{equation}
Num(t)=t(M-t+1)
\end{equation}
Consequently, different time indices are associated with different numbers of intervals. The number of intervals for each time index for time series with 1000 time points is plotted in Figure \ref{fig:intervalSample}. The indices in the middle have more intervals than the indices on the edges of the time series. Because $Imp_k(t)$ is calculated by adding the entropy gain of all the splits associated with time index $t$ for feature $f_k$, it can be biased towards the time points having more interval features (particularly if no feature is important for classification).

\def\myWidth{2.5}
\begin{figure*}[!]
\centering
\subfigure[The time series data and the importance curves from TSF.]{
\includegraphics[width=\myWidth in]{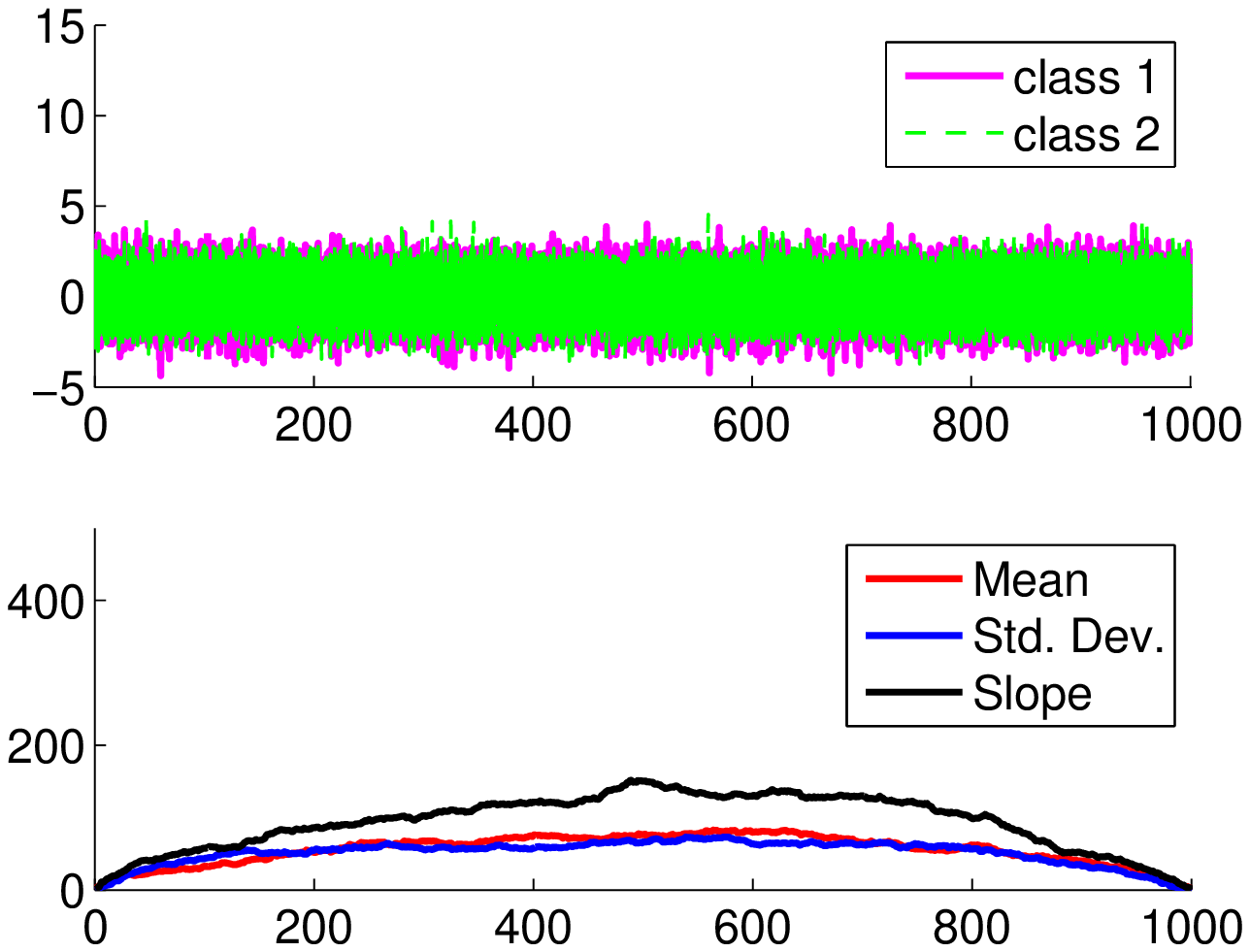}
\label{fig:simuUnifImpCurveEntrance}}
\subfigure[The number of intervals associated with each time index. The time indices in the middle are contained in more intervals.]{
\includegraphics[width=\myWidth in]{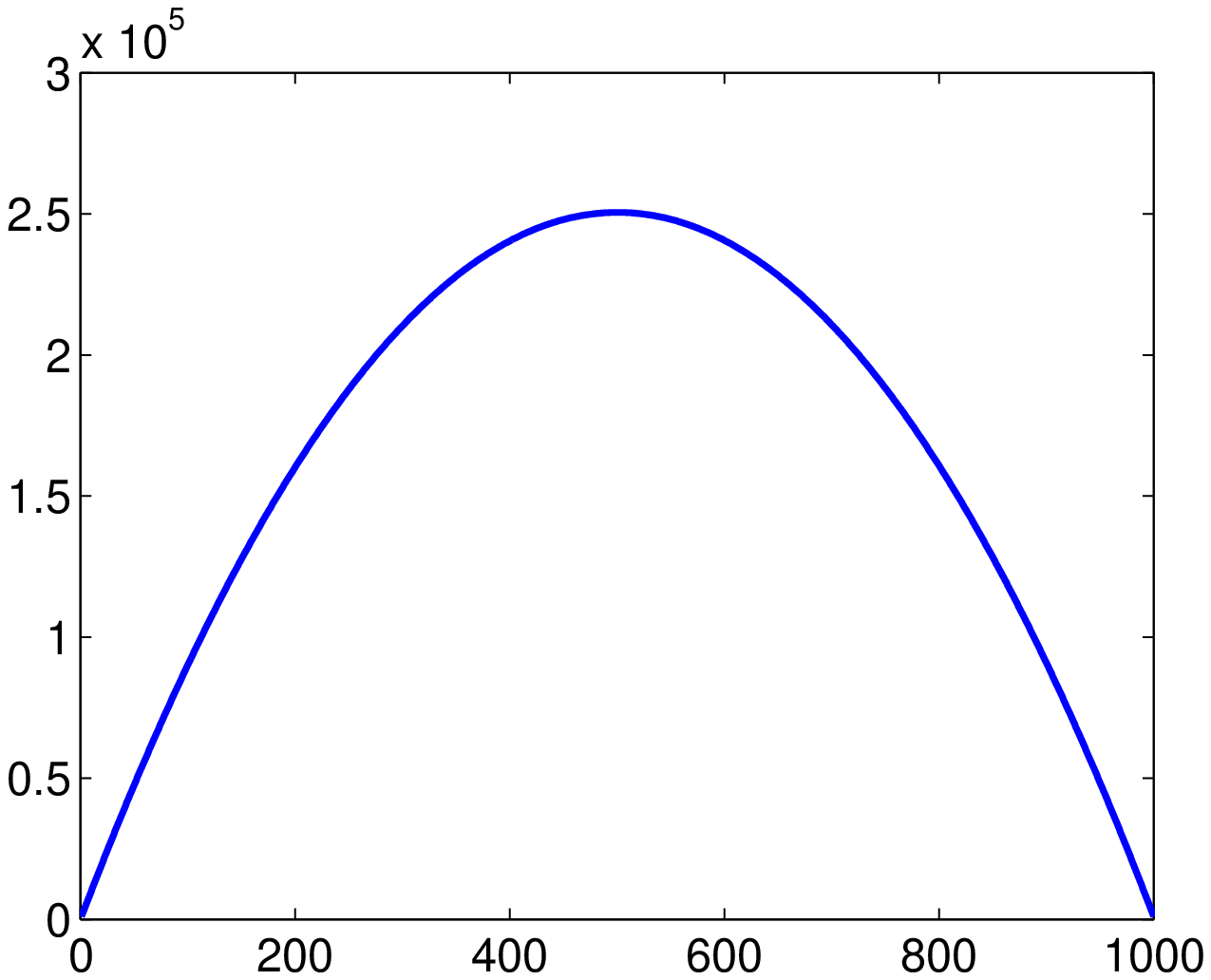}%\hfill
\label{fig:intervalSample}} %\hspace{0.1em}
\caption{When no feature is important for classification, the curves may be expected to have larger values for the middle indices as there are more intervals associated with the middle indices.}
\end{figure*}

\def\myWidth{2.5}
\def\myH{2}
\begin{figure*}[!]
\centering
\subfigure[The time series and the temporal importance curves obtained from TSF using Entrance gain.]{
\includegraphics[width=\myWidth in]{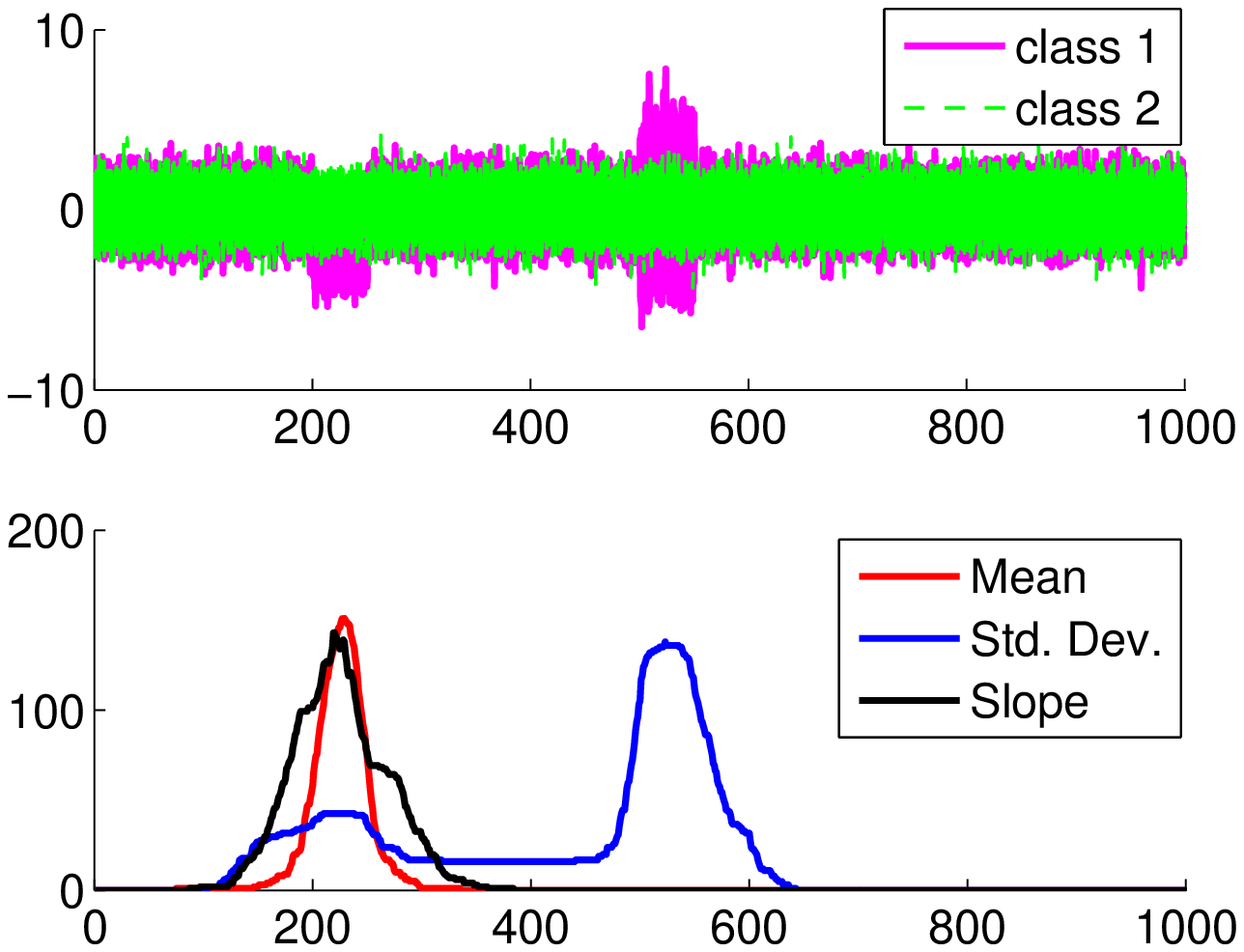}
\label{fig:simuImpCurveEntrance}}
\subfigure[The time series and the temporal importance curves obtained from TSF using entropy gain.]{
\includegraphics[width=\myWidth in]{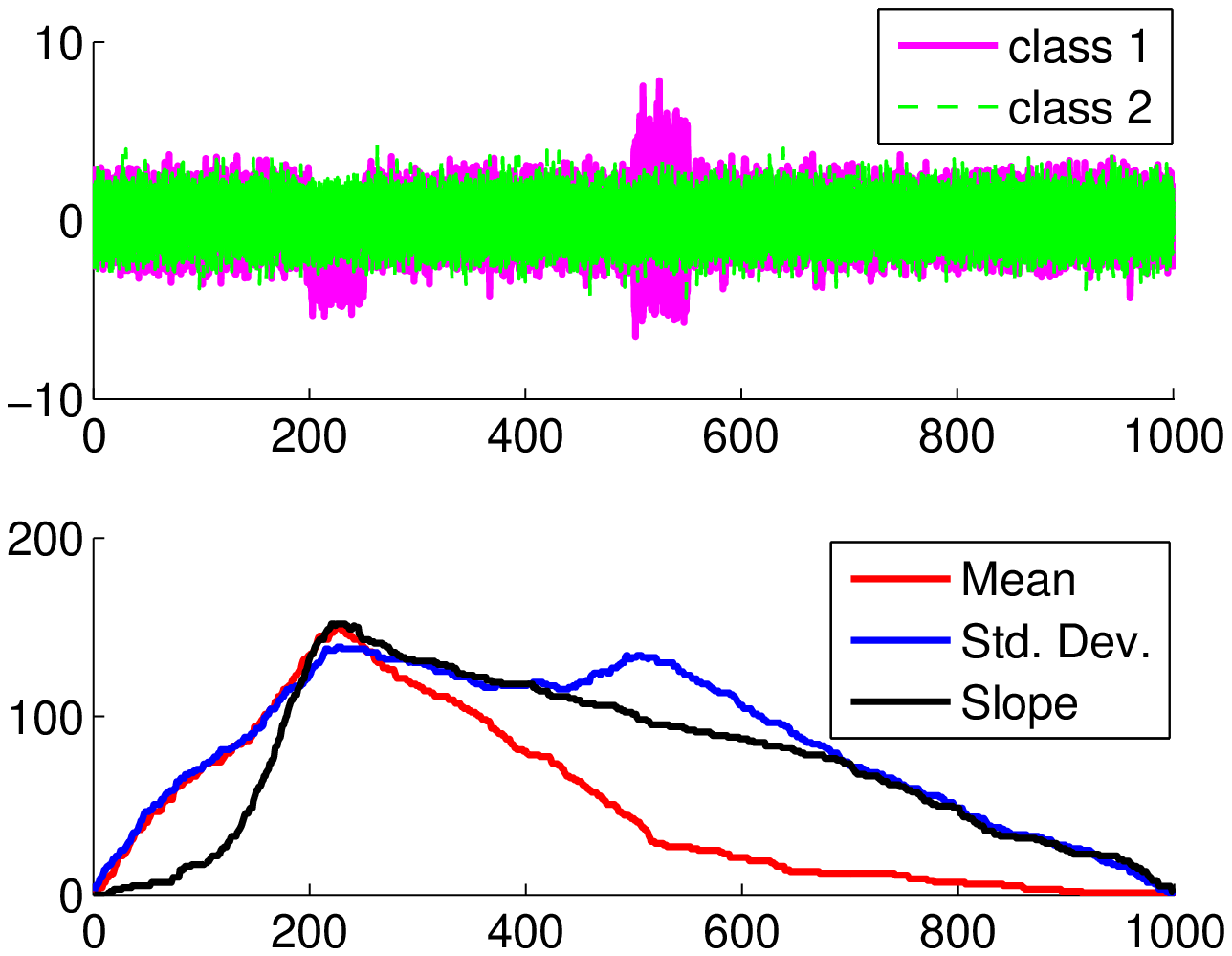}%\hfill
\label{fig:simuImpCurveEntrope}} %\hspace{0.1em}
\caption{The time series from the two classes differ in the mean in interval [201, 250], and differ in the standard deviation in interval [501, 550]. The importance curves from TSF using Entrance gain are able to capture the informative intervals well. The curves from TSF using entropy gain have peaks in interval [201, 250], but have long tails.}
\end{figure*}

For the second data set the time series from the two classes have different means in interval [201, 250], and different standard deviations in interval [501, 550]. The temporal importance curves from TSF using Entrance gain are shown in Figure \ref{fig:simuImpCurveEntrance}. The curves for the mean and slope have peaks in interval [201, 250], and the curve for the standard deviation has a peak in interval [501, 550]. Therefore, these curves capture the important temporal characteristics.

We also built TSF using entropy gain, and the corresponding temporal importance curves are shown in Figure \ref{fig:simuImpCurveEntrope}. Although the curves also have peaks in interval [201, 250], the curves have long tails. Indeed, the entropy gain is not able to distinguish many interval features. For example, the mean feature for interval [201,250], and the mean feature for interval [201,400] have the same entropy gain as both can distinguish the two classes of time series. However, the mean feature for interval [201,250] has a larger $E$ than the mean feature for interval [201,400]. Consequently, TSF using Entrance gain is able to capture the temporal characteristics more accurately.

\begin{table}[h!]
%\begin{sidewaystable}[!h]
\scriptsize
\centering
%\raggedleft
% Table generated by Excel2LaTeX from sheet 'New'
\begin{tabular}{|c|c|c|c|c|}
\hline
           &            &   Training &    Testing &            \\

           &     Length &  instances &  instances &    Classes \\
\hline
   50words &        270 &        450 &        455 &         50 \\
\hline
     Adiac &        176 &        390 &        391 &         37 \\
\hline
      Beef &        470 &         30 &         30 &          5 \\
\hline
       CBF &        128 &         30 &        900 &          3 \\
\hline
ChlorineConcentration &        166 &        467 &       3840 &          3 \\
\hline
CinC\_ECG\_torso &       1639 &         40 &       1380 &          4 \\
\hline
    Coffee &        286 &         28 &         28 &          2 \\
\hline
Cricket\_X &        300 &        390 &        390 &         12 \\
\hline
Cricket\_Y &        300 &        390 &        390 &         12 \\
\hline
Cricket\_Z &        300 &        390 &        390 &         12 \\
\hline
DiatomSizeReduction &        345 &         16 &        306 &          4 \\
\hline
    ECG200 &         96 &        100 &        100 &          2 \\
\hline
ECGFiveDays &        136 &         23 &        861 &          2 \\
\hline
   FaceAll &        131 &        560 &       1690 &         14 \\
\hline
  FaceFour &        350 &         24 &         88 &          4 \\
\hline
  FacesUCR &        131 &        200 &       2050 &         14 \\
\hline
      Fish &        463 &        175 &        175 &          7 \\
\hline
  GunPoint &        150 &         50 &        150 &          2 \\
\hline
   Haptics &       1092 &        155 &        308 &          5 \\
\hline
InlineSkate &       1882 &        100 &        550 &          7 \\
\hline
ItalyPowerDemand &         24 &         67 &       1029 &          2 \\
\hline
 Lighting2 &        637 &         60 &         61 &          2 \\
\hline
 Lighting7 &        319 &         70 &         73 &          7 \\
\hline
    MALLAT &       1024 &         55 &       2345 &          8 \\
\hline
MedicalImages &         99 &        381 &        760 &         10 \\
\hline
MoteStrain &         84 &         20 &       1252 &          2 \\
\hline
NonInvasiveFatalECG\_Thorax1 &        750 &       1800 &       1965 &         42 \\
\hline
NonInvasiveFatalECG\_Thorax2 &        750 &       1800 &       1965 &         42 \\
\hline
  OliveOil &        570 &         30 &         30 &          4 \\
\hline
   OSULeaf &        427 &        200 &        242 &          6 \\
\hline
SonyAIBORobotSurface &         70 &         20 &        601 &          2 \\
\hline
SonyAIBORobotSurfaceII &         65 &         27 &        953 &          2 \\
\hline
StarLightCurves &       1024 &       1000 &       8236 &          3 \\
\hline
SwedishLeaf &        128 &        500 &        625 &         15 \\
\hline
   Symbols &        398 &         25 &        995 &          6 \\
\hline
Syntheticcontrol &         60 &        300 &        300 &          6 \\
\hline
     Trace &        275 &        100 &        100 &          4 \\
\hline
TwoLeadECG &         82 &         23 &       1139 &          2 \\
\hline
TwoPatterns &        128 &       1000 &       4000 &          4 \\
\hline
uWaveGestureLibrary\_X &        315 &        896 &       3582 &          8 \\
\hline
uWaveGestureLibrary\_Y &        315 &        896 &       3582 &          8 \\
\hline
uWaveGestureLibrary\_Z &        315 &        896 &       3582 &          8 \\
\hline
     Wafer &        152 &       1000 &       6164 &          2 \\
\hline
WordsSynonyms &        270 &        267 &        638 &         25 \\
\hline
      Yoga &        426 &        300 &       3000 &          2 \\
\hline
\end{tabular}
\caption{Summary of the time series data sets: the number of training and testing instances, the number of classes and the lengths of the time series. \label{table:char}}
%\end{sidewaystable}
\end{table}

\begin{table}[h!]
\tiny
%\scriptsize
%\begin{sidewaystable}[!h]
\centering
\begin{tabular}{|c|c|c|c|c|c|c|}
\hline
    {\bf } &  {\bf TSF} &  {\bf TSF} &     {\bf } &   {\bf NN} &  {\bf DTW} &  {\bf DTW} \\

    {\bf } & {\bf Entrance} & {\bf entropy} & {\bf interRF} & {\bf Euclidean } & {\bf Best} & {\bf NoWin} \\
\hline
   50words &     0.2659 &     0.2769 &     0.2989 &     0.3690 &     0.2420 &     0.3100 \\
\hline
     Adiac &     0.2302 &     0.2609 &     0.2506 &     0.3890 &     0.3910 &     0.3960 \\
\hline
      Beef &     0.2333 &     0.3000 &     0.3000 &     0.4670 &     0.4670 &     0.5000 \\
\hline
       CBF &     0.0256 &     0.0389 &     0.0411 &     0.1480 &     0.0040 &     0.0030 \\
\hline
ChlorineConcentration &     0.2537 &     0.2596 &     0.2273 &     0.3500 &     0.3500 &     0.3520 \\
\hline
CinC\_ECG\_torso &     0.0391 &     0.0688 &     0.1065 &     0.1030 &     0.0700 &     0.3490 \\
\hline
    Coffee &     0.0357 &     0.0714 &     0.0000 &     0.2500 &     0.1790 &     0.1790 \\
\hline
Cricket\_X &     0.2897 &     0.2872 &     0.3128 &     0.4260 &     0.2360 &     0.2230 \\
\hline
Cricket\_Y &     0.2000 &     0.2000 &     0.2436 &     0.3560 &     0.1970 &     0.2080 \\
\hline
Cricket\_Z &     0.2436 &     0.2385 &     0.2436 &     0.3800 &     0.1800 &     0.2080 \\
\hline
DiatomSizeReduction &     0.0490 &     0.1013 &     0.0980 &     0.0650 &     0.0650 &     0.0330 \\
\hline
    ECG200 &     0.0800 &     0.0700 &     0.1700 &     0.1200 &     0.1200 &     0.2300 \\
\hline
ECGFiveDays &     0.0557 &     0.0697 &     0.1231 &     0.2030 &     0.2030 &     0.2320 \\
\hline
   FaceAll &     0.2325 &     0.2314 &     0.2497 &     0.2860 &     0.1920 &     0.1920 \\
\hline
  FaceFour &     0.0227 &     0.0341 &     0.0568 &     0.2160 &     0.1140 &     0.1700 \\
\hline
  FacesUCR &     0.1010 &     0.1088 &     0.1283 &     0.2310 &     0.0880 &     0.0951 \\
\hline
      Fish &     0.1543 &     0.1543 &     0.1486 &     0.2170 &     0.1600 &     0.1670 \\
\hline
  GunPoint &     0.0467 &     0.0467 &     0.0400 &     0.0870 &     0.0870 &     0.0930 \\
\hline
   Haptics &     0.5520 &     0.5649 &     0.5487 &     0.6300 &     0.5880 &     0.6230 \\
\hline
InlineSkate &     0.6818 &     0.6746 &     0.6873 &     0.6580 &     0.6130 &     0.6160 \\
\hline
ItalyPowerDemand &     0.0301 &     0.0330 &     0.0321 &     0.0450 &     0.0450 &     0.0500 \\
\hline
 Lighting2 &     0.1803 &     0.1803 &     0.2459 &     0.2460 &     0.1310 &     0.1310 \\
\hline
 Lighting7 &     0.2603 &     0.2603 &     0.2740 &     0.4250 &     0.2880 &     0.2740 \\
\hline
    MALLAT &     0.0448 &     0.0716 &     0.0644 &     0.0860 &     0.0860 &     0.0660 \\
\hline
MedicalImages &     0.2237 &     0.2316 &     0.2658 &     0.3160 &     0.2530 &     0.2630 \\
\hline
MoteStrain &     0.1190 &     0.1182 &     0.0942 &     0.1210 &     0.1340 &     0.1650 \\
\hline
NonInvasiveFatalECG\_Thorax1 &     0.0987 &     0.1033 &     0.1104 &     0.1710 &     0.1850 &     0.2090 \\
\hline
NonInvasiveFatalECG\_Thorax2 &     0.0865 &     0.0936 &     0.0875 &     0.1200 &     0.1290 &     0.1350 \\
\hline
  OliveOil &     0.0667 &     0.1000 &     0.1333 &     0.1330 &     0.1670 &     0.1330 \\
\hline
   OSULeaf &     0.4339 &     0.4256 &     0.4587 &     0.4830 &     0.3840 &     0.4090 \\
\hline
SonyAIBORobotSurface &     0.2330 &     0.2346 &     0.2562 &     0.1410 &     0.1410 &     0.1690 \\
\hline
SonyAIBORobotSurfaceII &     0.1868 &     0.1773 &     0.2067 &     0.3050 &     0.3050 &     0.2750 \\
\hline
StarLightCurves &     0.0357 &     0.0364 &     0.0327 &     0.1510 &     0.0950 &     0.0930 \\
\hline
SwedishLeaf &     0.1056 &     0.1088 &     0.0768 &     0.2130 &     0.1570 &     0.2100 \\
\hline
   Symbols &     0.1116 &     0.1206 &     0.1216 &     0.1000 &     0.0620 &     0.0500 \\
\hline
Syntheticcontrol &     0.0267 &     0.0233 &     0.0167 &     0.1200 &     0.0170 &     0.0070 \\
\hline
     Trace &     0.0200 &     0.0000 &     0.0400 &     0.2400 &     0.0100 &     0.0000 \\
\hline
TwoLeadECG &     0.1177 &     0.1115 &     0.1773 &     0.2530 &     0.1320 &     0.0960 \\
\hline
TwoPatterns &     0.0543 &     0.0530 &     0.0153 &     0.0900 &     0.0015 &     0.0000 \\
\hline
uWaveGestureLibrary\_X &     0.2102 &     0.2127 &     0.2094 &     0.2610 &     0.2270 &     0.2730 \\
\hline
uWaveGestureLibrary\_Y &     0.2876 &     0.2881 &     0.3023 &     0.3380 &     0.3010 &     0.3660 \\
\hline
uWaveGestureLibrary\_Z &     0.2624 &     0.2669 &     0.2764 &     0.3500 &     0.3220 &     0.3420 \\
\hline
     Wafer &     0.0054 &     0.0047 &     0.0071 &     0.0050 &     0.0050 &     0.0200 \\
\hline
WordsSynonyms &     0.3793 &     0.3809 &     0.4138 &     0.3820 &     0.2520 &     0.3510 \\
\hline
      Yoga &     0.1513 &     0.1567 &     0.1380 &     0.1700 &     0.1550 &     0.1640 \\
\hline
{\bf win/lose/tie} &    {\bf -} & {\bf 16/28/1} & {\bf 13/32/0} & {\bf 4/41/0} & {\bf 17/28/0} & {\bf 16/29/0} \\
\hline
{\bf Average rank} & {\bf 2.48} & {\bf 2.86} & {\bf 3.43} & {\bf 5.04} & {\bf 3.31} & {\bf 3.88} \\
\hline
{\bf Rank difference} &    {\bf -} & {\bf 0.38} & {\bf 0.96} & {\bf 2.57} & {\bf 0.83} & {\bf 1.40} \\
\hline
{\bf Wilcoxon} &    {\bf -} & {\bf 0.007} & {\bf 0.000} & {\bf 0.000} & {\bf 0.065} & {\bf 0.006} \\
\hline
\end{tabular}
\caption{{The error rates of TSF using the splitting criterion: Entrance gain (TSF) or entropy gain (TSF-entropy), random forest with 500 trees applied to the interval features with sizes power of two (interRF), 1-NN with Euclidean distance (NNEuclidean), 1-NN with the best warping window DTW (DTWBest) \citep{ratanamahatana2004making}, and 1-NN DTW with no warping window (DTWNoWin). The win-lose-tie results of each competitor compared to TSF, the average rank of each classifier, the rank difference and the Wilcoxon signed ranks test between TSF and each competitor are also calculated. When multiple methods have the same error rate for a data set, the average rank is used. For example, both DTWBest and DTWNoWin have the minimum error rate 0.192 for the FaceAll data set, and, thus, the rank for both is 1.5.}\label{table:err}}
%\end{sidewaystable}
\end{table}

\section{Experiments \label{sec:exp}}
\subsection{Experimental Setup}
The main functions of the TSF algorithm were implemented in Matlab, while computationally expensive subfunctions such as interval feature calculations were written in C. The parameters were set as follows: the number of trees = 500, $f(\cdot)=\{mean, standard\ deviation, slope\}$, and the number of candidate thresholds $\kappa=20$. TSF was applied to a set of time series benchmark data sets \citep{keogh2008ucr} summarized in Table \ref{table:char}. The training/testing split setting is the same as in \citet{keogh2008ucr}. The experiments were run on a computer with four cores and the TSF algorithm was built in parallel. %multiple cores were used for building TSF in parallel.

The purpose of the experiments is to answer the following questions: (1) Does the Entrance gain criterion improve the accuracy performance and how is the accuracy performance of TSF compared to other time series classifiers? (2) Is TSF computationally efficient? (3) Can the temporal importance curves provide some insights about the temporal characteristics useful for classification?

\subsection{Results}
We investigated the performance of TSF using the Entrance gain criterion (denoted as TSF) and using the original entropy gain criterion (denoted as TSF-entropy), respectively. We also considered alternative classifiers for comparison: random forest \cite{breiman2001} applied %to the original time series points (origRF), and
to the interval features with sizes power of two (interRF), the 1-nearest-neighbor (NN) classifier with Euclidean distance (NNEuclidean), the 1-NN Best warping window DTW (DTWBest) \citep{ratanamahatana2004making} and the 1-NN DTW with no warping window (DTWNoWin) methods acquired directly from \citet{keogh2008ucr}. DTWBest has a fixed window limiting the window width and searches for the best window size, while DTWNoWin does not use such a window. %The ``randomForest" R package \citep{Andy2002} was used for running random forest with the default setting.

The classification error rates are shown in Table \ref{table:err}. To compare multiple classifiers to TSF over multiple data sets, we used the procedure for comparing multiple classifiers with a control over multiple data sets suggested by \citet{demvsar2006statistical}, i.e., the Friedman test \cite{friedman1940comparison} followed by the Bonferroni-Dunn test \cite{dunn1961multiple} if the Friedman test shows a significant difference between the classifiers. In our case, the Friedman test shows that there is a significant difference between the six classifiers at the 0.001 level. Therefore, we proceeded with the Bonferroni-Dunn test.

For the Bonferroni-Dunn test, the performance of two classifiers is different at the $\alpha$ level if the their average ranks differ by at least the critical difference (CD):
\begin{equation}\label{eq:BDtest}
z_\alpha=q_\alpha\sqrt{\frac{N_{classifier}(N_{classifier}+1)}{6N_{data}}}
\end{equation}
where $N_{classifier}$ is the number of classifiers in the comparison (six classifiers in our experiments), $N_{data}$ is the number of data sets (45 data sets in our experiments), and $q_\alpha$ is the critical value for the two-tailed Bonferroni-Dunn test for multiple classifier comparison with a control. Note $q_{0.05}=2.576$ and $q_{0.1}=2.326$ (Table 5(b) in \citet{demvsar2006statistical}), then according to Equation \ref{eq:BDtest}, $z_{0.05}=1.016$ and $z_{0.1}=0.917$. The average rank of each classifier, and the difference between the average ranks of TSF and each competitor are shown in Table \ref{table:err}. According to the rank difference, there is a significant difference between TSF and competitors NNEuclidean, DTWNoWin and interRF at the 0.1 level.

In addition to the multi-classifier comparison procedure, we also considered Wilcoxon signed ranks test \cite{wilcoxon1945individual} suggested for comparing a pair of classifiers, as the resolution for the multi-classifier comparison procedure can be too low to distinguish two classifiers with significantly different performance, but with close average ranks. For example, for six classifiers and 45 data sets, assume classifier A always ranks the first and classifier B always ranks the second. Although classifier A is always better than classifier B, the average ranks of classifier A and classifier B differ by only one, and therefore there is no significant difference between the two classifiers at the 0.05 level according to the two-tailed Bonferroni-Dunn test.

The p-values of the Wilcoxon signed ranks tests between TSF and each competitor are shown in Table \ref{table:err}. It can be seen there is a significant difference between TSF and all other competitors: TSF-entropy, interRF, NNEuclidean, DTWNoWin and DTWBest at the 0.1 level. %Therefore, the Entrance gain criterion can improve the accuracy performance of TSF, and TSF outperforms other .

\def\myWidth{2.5}
\begin{figure}[!h]
\centering
    \includegraphics[width=4 in,height=3 in]{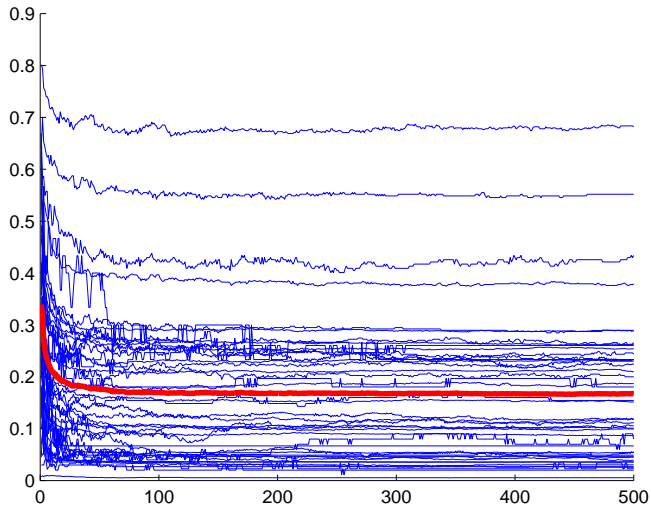}
\caption{Plot of the error rate of each data set versus the number of trees in TSF, and the average error rate over all data sets versus the number of trees (represented by the thicker red line). We want to show the trend so different data sets are not distinguished. The error rates tend to decrease as the number of trees increases, but the change is relatively small for most data sets after 100 trees.\label{fig:converge}}
\end{figure}

Next consider the robustness of TSF accuracy to the number of trees. Figure \ref{fig:converge} shows the error rate of each data set versus the number of trees, and the average error rate over all data sets versus the number of trees (represented by the thicker red line). The error rates tend to decrease as the number of trees increases, but the change is relatively small for most data sets after 100 trees.

\def\myWidth{2.5}
\def\myH{2}
\begin{figure*}[!]
\centering
\subfigure[GunPoint time series]{
\includegraphics[width=\myWidth in]{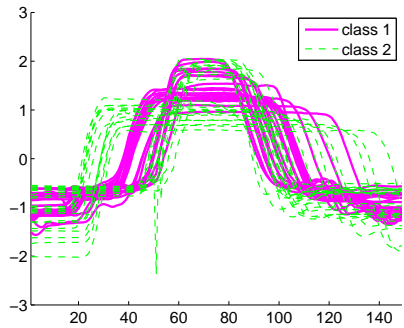}%\hfill
} %\hspace{0.1em}
\subfigure[Wafer time series]{
\includegraphics[width=\myWidth in]{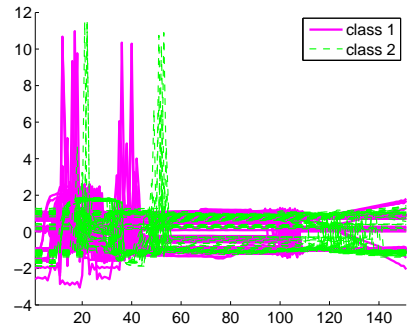}%\hfill
} %\hspace{0.1em}
\subfigure[The temporal importance curves for the GunPoint data.]{
\includegraphics[width=\myWidth in]{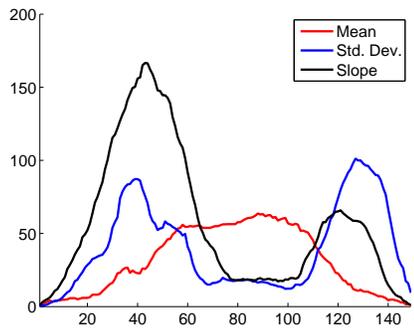}%\hfill
} %\hspace{0.1em}
\subfigure[The temporal importance curves for the Wafer data.]{
\includegraphics[width=\myWidth in]{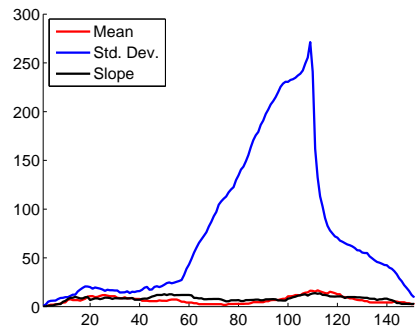}
}
\caption{The time series and the temporal importance curves (mean, standard deviation and slope) for the GunPoint data set and the Wafer data set, respectively. \label{fig:importanceData}}
\end{figure*}

The GunPoint and Wafter time series and their corresponding temporal importance curves (mean, standard deviation and slope) are shown in Figure \ref{fig:importanceData}. For the GunPoint time series, the mean temporal importance curve captures the characteristic that the two classes have different means in interval [60,100]. The standard deviation and slope temporal importance curves, respectively, capture the characteristics that the two classes have different standard deviations and slopes in the left and right sides of the time series. For the Wafer time series, the standard deviation temporal importance curve captures the sudden changes of the time series of class 1 near the $100^{th}$ point. Consequently, the temporal importance curve is able to provide insights into the temporal characteristics useful for distinguishing time series from different classes.

\subsection{Computational Complexity}
First consider the computational complexity of TSF with regard to the length of time series. We selected the data sets with more than 1000 time points. For each data set, $\lambda M$ of the time points were randomly sampled, where $M$ is the length of the time series, and $\lambda$ is a multiplier. The computational times for different values of $\lambda$ are shown in Figure \ref{fig:timeFea}. Next consider the computational complexity of TSF with regard to the number of training instances. Data sets with more than 1000 training instances were selected. For each data set, $\lambda N$ of the time points were randomly sampled, where $N$ is the number of training instances. The computational times for different values of $\lambda$ are shown in Figure \ref{fig:timeExample}. It can be seen that the computational time tends to be linear both in the time series length and in the number of training instances.

Therefore, TSF is a computationally efficient classifier for time series. Furthermore, in the current TSF implementation, the interval features are dynamically calculated at each node, as pre-computing the interval features would need $O(M^2)$ features to be stored. It should noted, however, dynamic calculation can lead to repeated calculations of the interval features. %Experimental studies show that, the number of repeated calculations tend to be higher for data sets with a larger number of instances, maybe due to larger trees.
Therefore, the implementation can be further improved by storing the interval features already calculated to avoid repeated calculations.

\def\myWidth{2.5}
\begin{figure*}[!]
\centering
\subfigure[The computational time of TSF with regard to the time series length]{
\includegraphics[width=\myWidth in]{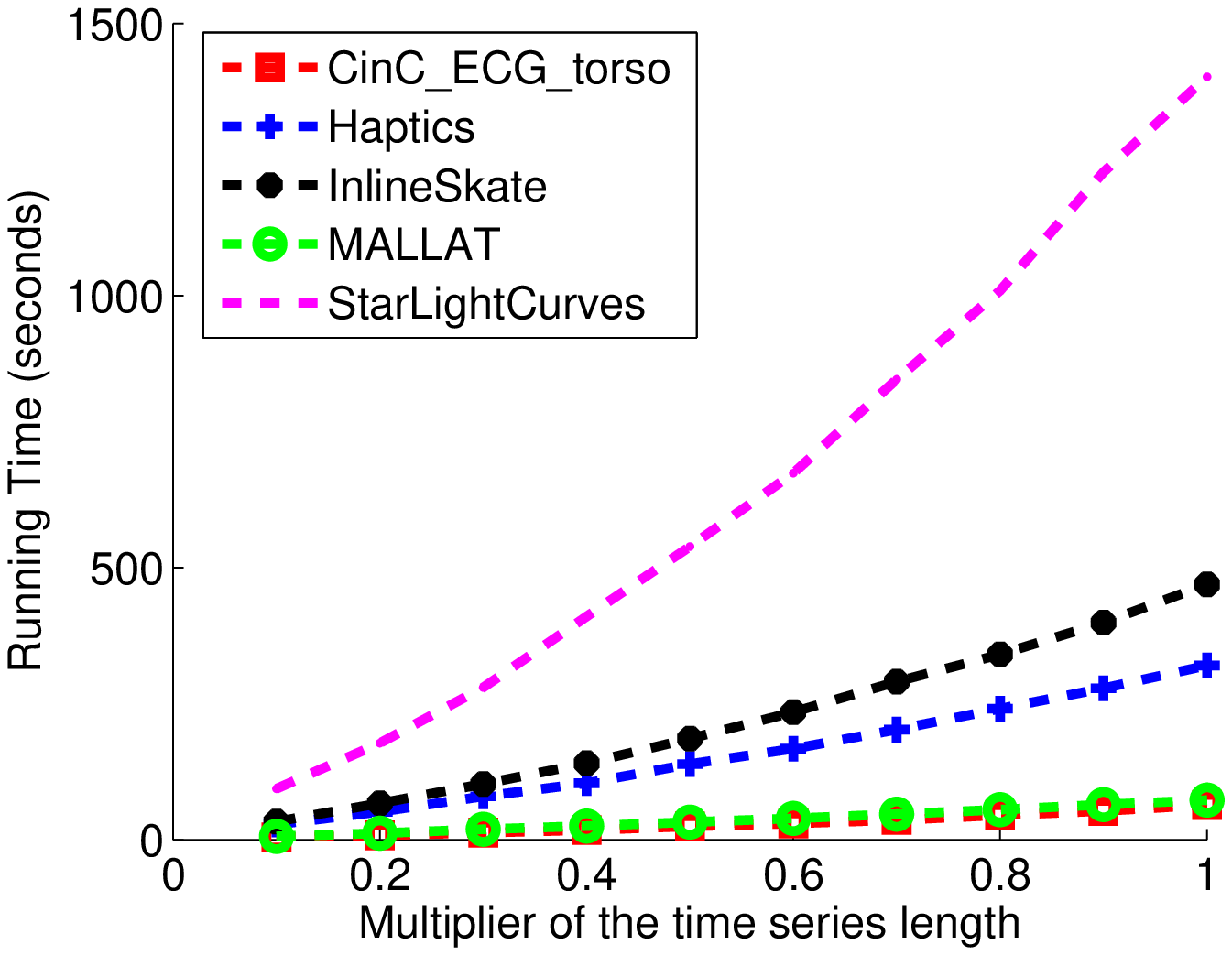}
\label{fig:timeFea}}
\subfigure[The computational time of TSF with regard to the number of training instances]{
\includegraphics[width=\myWidth in]{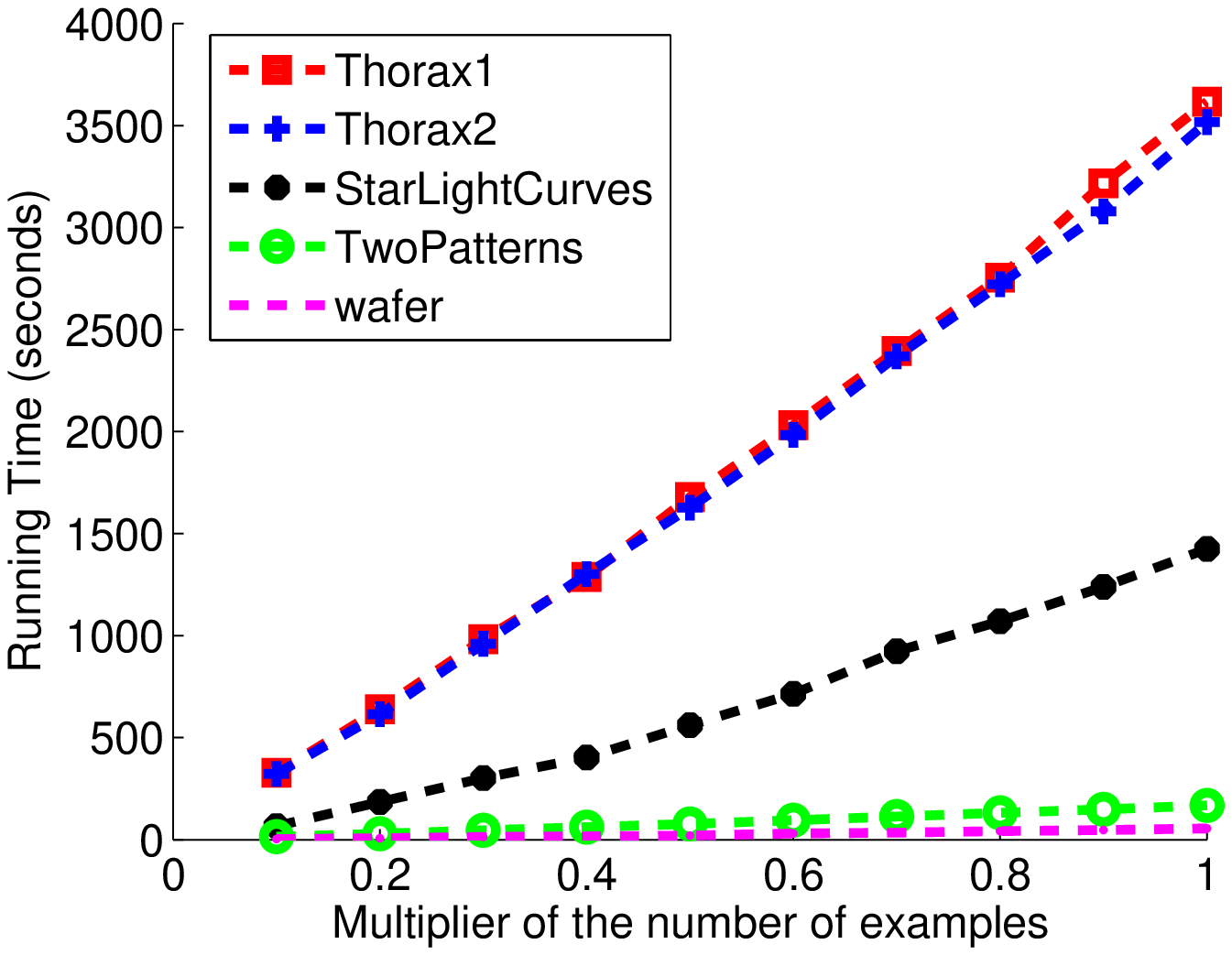}
\label{fig:timeExample}}

\caption{The computational time of TSF with regard to the time series length and the number of training instances, respectively. Data sets with more than 1000 time points and 1000 training instances were selected, respectively. The computational time tends to be linear both in the time series length and in the number of training instances.  \label{fig:computationalcomplexity}}
\end{figure*}

\section{Conclusions}\label{sec:conclusions}
Both high accuracy and interpretability are desirable for classifiers. Previous classifiers such as NNDTW can be accurate, but provide limited insights into the temporal characteristics. Interval features can be used to capture temporal characteristics, however, the huge feature space can result in many splits having the same entropy gain. Furthermore, the computational complexity becomes a concern when the feature space becomes large.

Time series forest (TSF) proposed here addresses the challenges by using the following two strategies. Firstly, TSF uses a new splitting criterion named Entrance gain that combines the entropy gain and a distance measure to identify high-quality splits. Experimental studies on 45 benchmark data sets show that the Entrance gain improves the accuracy of TSF. Secondly, TSF randomly samples $O(M)$ features from $O(M^2)$ features, and thus makes the computational complexity linear in the time series length. In addition, each tree in TSF is grown independently, and, therefore, modern parallel computing techniques can be leveraged to speed up TSF.

TSF is an ensemble of trees and is not easy to understand. However, we propose the temporal importance curve, calculated from TSF, to capture the informative interval features. The temporal importance curve enables one to identify the important temporal characteristics.

TSF uses simple summary statistical features, but outperforms widely used alternatives. %such as the nearest-neighbor classifiers with dynamic time warping.
More complex features, such as wavelets, can be also used in the framework of TSF, which potentially can further improve the accuracy performance, but at the cost of interpretability.

In summary, TSF is an accurate, efficient time series classifier, and is able to provide insights on the temporal characteristics useful for distinguishing time series from different classes. We also note that TSF assumes that the time series are of the same length. Given a set of time series with different lengths, techniques such as dynamic time warping can be used to align the time series into the same length. Still, directly handling time series with varying lengths would make TSF more convenient to use, and future work includes such an extension.% Consequently, future work includes extending TSF to time series with different lengths.

\section*{Acknowledgements}
This research was partially supported by ONR grant N00014-09-1-0656. We also wish to thank the editor and anonymous reviewers for their valuable comments.

\bibliographystyle{model1b-num-names}%splncs03 elsarticle-num  harv \bibliographystyle{alpha}  model1b-num-names.bst  model1-num-names
\bibliography{timeF}

\end{document}